# EvoCLINICAL: Evolving Cyber-Cyber Digital Twin with Active Transfer Learning for Automated Cancer Registry System

Chengjie Lu
Simula Research Laboratory and University of Oslo
Oslo, Norway
chengjielu@simula.no

Qinghua Xu
Simula Research Laboratory and University of Oslo
Oslo, Norway
qinghua@simula.no

Tao Yue
Simula Research Laboratory
Oslo, Norway
tao@simula.no

Shaukat Ali
Simula Research Laboratory and Oslo Metropolitan University
Oslo, Norway
shaukat@simula.no

Thomas Schwitalla
Cancer Registry of Norway
Oslo, Norway
thsc@kreftregisteret.no

Jan F. Nygård
Cancer Registry of Norway
Oslo, Norway
UiT The Arctic University of Norway
Tromsø, Norway
jfn@kreftregisteret.no

## ABSTRACT

The Cancer Registry of Norway (CRN) collects information on cancer patients by receiving cancer messages from different medical entities (e.g., medical labs, hospitals) in Norway. Such messages are validated by an automated cancer registry system: GURI. Its correct operation is crucial since it lays the foundation for cancer research and provides critical cancer-related statistics to its stakeholders. Constructing a cyber-cyber digital twin (CCDT) for GURI can facilitate various experiments and advanced analyses of the operational state of GURI without requiring intensive interactions with the real system. However, GURI constantly evolves due to novel medical diagnostics and treatment, technological advances, etc. Accordingly, CCDT should evolve as well to synchronize with GURI. A key challenge of achieving such synchronization is that evolving CCDT needs abundant data labelled by the new GURI.

To tackle this challenge, we propose EvoCLINICAL, which considers the CCDT developed for the previous version of GURI as the pretrained model and fine-tunes it with the dataset labelled by querying a new GURI version. EvoCLINICAL employs a genetic algorithm to select an optimal subset of cancer messages from a candidate dataset and query GURI with it. We evaluate EvoCLINICAL on three evolution processes. The precision, recall, and F1 score are all greater than 91%, demonstrating the effectiveness of EvoCLINICAL. Furthermore, we replace the active learning part of EvoCLINICAL with random selection to study the contribution of transfer learning to the overall performance of EvoCLINICAL. Results show that employing active learning in EvoCLINICAL increases its performance consistently.

## 1 INTRODUCTION

Cancer is a disease that affects numerous individuals in the world. It is reported that the number of new cancer cases is around 20 million, and almost 10 million people died from cancer in 2020 only [16]. Cancer inflicts pain on a patient not only physically and mentally but also significantly impacts society as a whole.

The Cancer Registry of Norway (CRN) collects cancer data and provides statistics and research data to the public, government, hospitals and researchers. CRN possesses a rich dataset on cancer patients since health professionals in Norway are instructed by law to notify CRN of diagnostics, treatment and follow-up of cancer patients. Each cancer case is aggregated using a set of cancer messages containing specific information on medical events. To ensure the high quality of the data, CRN developed an Automated Cancer Registry System (known as GURI) to check the message's validity of a set of validation rules defined based on standard medical rules (e.g., a person with type O blood cannot have a biological parent with type AB blood). Specifically, GURI is constructed with domain knowledge and maintains a set of validation rules. The validity of each message is determined by systematically checking violations of these rules. The correct operation of GURI is the foundation of producing reliable data for facilitating various research activities. SSimulation, which imitates a real-world system with software, is one promising approach to studying the operation of GURI. Interaction with simulation instead of the real system is comparatively cheaper and more malleable, i.e., it facilitates various experiments without intervening too much with the real system. Researchers have recently proposed extending the simulation concept into a digital twin (DT), synchronising the operation state with the real system. Such an extension enables DT to uncover unknown issues before they cause irreversible damage to the real system. Moreover, it is possible to employ DT to replay critical events and assess how to mitigate them in the future.

Despite their great success in various domains [15], most existing DTs are confined to cyber-physical digital twins, where DTs are used to simulate physical systems. Nonetheless, we posit that critical software systems like GURI can also benefit from DT. Facebook proposed a cyber-cyber digital twin (CCDT) for one of their platforms [1]. We take inspiration from this work and build CCDT for GURI. With CCDT, we expect to enable the simulation of socio-technical behaviours of various stakeholders of GURI, including medical coders, hospital professionals, and policymakers, who constantly interact with GURI and consequently impact its operation. This is important to develop a test infrastructure for GURI. However, GURI keeps evolving naturally as novel medical diagnostics and treatments are constantly introduced, new medical knowledge is gained, etc., which entails adding, deleting, and modifying the validation rules in GURI. Consequently, CCDT has to be continuously evolved to remain synchronized with evolving GURI. Otherwise, the performance of CCDT would present an unexpected decrease inevitably. One intuitive evolution method is to train a new model from scratch with abundant labelled new data. However, such labels are acquired through interaction with GURI, which can cause detriment to its safety and security. Therefore, we aspire to evolve CCDT with a small amount of labelled data.

To that end, we propose a novel method EvoCLINICAL, which constructs a neural network-based CCDT for GURI and harnesses active transfer learning (TL) to enable the evolution of CCDT. Specifically, CCDT simulates a specific version of GURI by predicting the validation result of a given cancer message. Consequently, a well-trained CCDT possesses knowledge about how a specific version of GURI validates cancer messages. Such knowledge can be leveraged to build a CCDT for a new version of GURI. Therefore, when a major version update is conducted, we employ TL to evolve the CCDT, which entails a pretraining + fine-tuning paradigm. TL treats the CCDT from the previous version as a pretrained model and fine-tunes it with data labelled by the updated GURI. However, fine-tuning necessitates sufficient data labelled by the updated GURI, whereas minimal interaction with GURI is preferred. As a solution, we adopt an active learning paradigm and select a subset of the most valuable cancer messages from a candidate dataset to query the new GURI. These labelled cancer messages are then used to fine-tune the CCDT, synchronizing it with the new GURI.

To evaluate EvoCLINICAL, we acquire six versions of GURI and a candidate dataset. The candidate dataset is generated following the specifications of a cancer message, containing 8000 unlabelled cancer messages. These six versions are divided into three experimental groups, with each group consisting of a source GURI and a target GURI. We perform separate TL from source GURI to target GURI in each group. Experimental results show EvoCLINICAL's high performance in terms of precision, recall and F1 score, with a minimum of 0.9105 (i.e., recall value on $S_2 \rightarrow T_2$). We also find active learning improves the performance in all three evolution processes compared to random selection. Additional experiments show large candidate dataset sizes have positive influences on EvoCLINICAL's performance, as expected.

The contribution of EvoCLINICAL is as follows. 1) We propose a neural network-based CCDT that simulates GURI for validating cancer messages. In particular, we first extract categorical, numerical, and string features from cancer messages, which are fed into a Convolutional Neural Network (CNN) to predict the results for each rule. 2) We leverage TL to evolve the CCDT for the new version of GURI. TL adopts a pretraining + fine-tuning paradigm, harnessing knowledge from previous versions of GURI to build a CCDT for the new version of GURI. 3) To reduce the query times on the new version of GURI, we propose a novel active learning method to select the most valuable cancer messages for GURI to label. Active learning utilizes indicator-based evolutionary algorithm IBEA [45] to select a subset from the candidate dataset with five optimization objectives. 4) We evaluate EvoCLINICAL with six versions of GURI. We first assess the effectiveness of TL and active learning, then analyse the impact of the candidate dataset size on the performance of EvoCLINICAL.

The rest of the paper is organized as follows. We first present the real-world application context in Section 2. In Section 3, we demonstrate the details of EvoCLINICAL, including CCDT construction and evolution. Section 4 presents the experiment design of EvoCLINICAL, followed by the experiment results in Section 5. In Section 6, we discuss the practical implications of EvoCLINICAL and present the lessons learned. Related works are listed in Section 7, and we identify potential threats to the validity of the empirical study in Section 8. Finally, we conclude this paper and discuss future works in Section 9.

## 2 REAL-WORLD APPLICATION CONTEXT AND CHALLENGES

CRN is one of the oldest national cancer registries in the world, extensively researching cancer since 1951. Over the years, the practices of managing cancer registries in CRN have shifted from a manual approach to a rule-based system, i.e., GURI. CRN collects cancer messages from different medical entities throughout Norway and relies on GURI to assess the validity of these messages. We present a snippet of one cancer message as in Listing 1. A cancer message is created as a JSON-like file, encompassing multiple fields related to the cancer registry. These fields can be divided into three types: categorical fields (e.g., "gender", "topography" and "morphology" ), numerical fields (e.g., "chemotherapy"), and textual fields (e.g., "birth_date", "diagnosis_date", "ct" and "message_version"). GURI validates such a message by checking violations of the validation rules, each involving one or multiple fields in the message. Take this validation rule as an example: *The birth date of a patient shall be later than the diagnosis date*. It involves two fields in the message, i.e., "birth_date" and "diagnosis_date." Formally, let $X$ be a cancer message and $n$ be the rule set size of GURI. GURI validate $X$ by checking the violations of each rule $R^i$ and assign a label $y^i \in$ {"info", "warning", "not applied", "error"} to it, with "info" denoting that $X$ successfully passes $R^i$'s validation, "error" representing a failed validation result, "warning" highlighting improbable field values or combination values in $X$, e.g., the age of one patient is 130 years, and "not applied" meaning $R^i$ does not apply to $X$ due to unmet prerequisites, e.g., rule $R^i$ shall be activated on 2023/01/01 while $X$ is generated on 2020/01/01.

```
{
    "gender":"M",
    "topography": "809",
    "morphology" : "405",
    "chemotherapy": 3,
    "birth_date" : "2000-01-01",
    "diagnosis_date":"2019-07-09",
    "ct": "gvEQyqbV46",
    "message_version":"tNJP2eAMEd",
}
```

Listing 1: Snippet of a cancer message

The relations between fields in modern medical research are constantly changing, and new relations are being identified as research progresses. This is due to the development of advanced medical technologies that allow researchers to gain a more comprehensive picture of health than ever before. For example, genomics, cancer blood tests, MRIs, sleep analysis, and many other innovations allow medical professionals to gain the most comprehensive picture of patient health. Consequently, GURI needs to evolve by adding new rules or modifying existing rules when needed. From CRN, we obtained six stable versions of GURI. We divide them into three evolution processes for evaluation purposes as shown in Table 1. We can observe that the first version change $S_1 \rightarrow T_1$ concerns rules checking individual fields, whereas $S_2 \rightarrow T_2$ and $S_3 \rightarrow T_3$ introduce new rules checking more than one field. Concretely, $S_2 \rightarrow T_2$ introduces two new rules that involve two fields, and $S_3 \rightarrow T_3$ presents a new rule that involves three fields: diagnosis date, morphology, and topography. Evidently, the complexity of the newly introduced rules varies in the three evolution processes ($Complexity(S_1 \rightarrow T_1) < Complexity(S_2 \rightarrow T_2) < Complexity(S_3 \rightarrow T_3)$).

## 3 METHODOLOGY

We aim to build a CCDT for each source GURI and evolve it to synchronize with the corresponding target GURI. Concretely, EvoCLINICAL tackles the following challenges: The first challenge is about cancer message complexity. Despite its structural format, each cancer message contains multiple heterogeneous fields, such as topography type (categorical), chemotherapy times (numerical), and CT description (textual). Extracting high-quality features from these fields is non-trivial. As a solution, we leverage one-hot encoding, normalization, and universal sentence encoder to preprocess each field and harness a multi-layer CNN to extract local and global features from each message.

| Evolution Pair | $n_S$ | $n_T$ | Description of Changed Rules |
|---|---|---|---|
| $S_1 \rightarrow T_1$ | 30 | 35 | Rules related to non-solid surgery procedure, diagnosis date, and metastasis morphology. |
| $S_2 \rightarrow T_2$ | 40 | 45 | Rules related to clinical stage, tumour topography, combination validity of surgery type and topography, combination validity of morphology and basis. |
| $S_3 \rightarrow T_3$ | 51 | 56 | Rules related to melanoma, disease stage, combination validity of topography and tumour basis, combination validity of morphology, topography, and diagnosis date, and combination validity of morphology and tumour basis. |

Table 1: Descriptions of evolution processes of GURI. $n_S$ and $n_T$ denote the rule set sizes in the source and target GURI.

Another challenge is the data scarcity issue on target GURI. Constructing CCDT necessitates sufficient version-specific labelled data provided by GURI. For the source versions of GURI: $S_1$, $S_2$, and $S_3$, we hypothesize that they produce abundant labelled data due to their continuous operation in the production environment. By contrast, the number of labelled data from the target versions, i.e., $T_1$, $T_2$, and $T_3$, is not guaranteed due to their short operation time. Meanwhile, querying the target GURI with a large number of unlabeled messages should be avoided since it can interfere with the normal operation of the target GURI. To mitigate this problem, we postulate that the old CCDT for source versions contains useful knowledge for constructing CCDT for target versions since not all rules are updated, and these rules are interdependent on each other. Therefore, we harness TL to transfer such knowledge from source CCDTs to target CCDTs, theoretically requiring less data than training from scratch. TL considers the source CCDT as a pretrained model for the target version and fine-tunes it with data acquired from the target version.

Despite that it requires few data than training from scratch, acquiring sufficient data from target GURI for fine-tuning is still challenging. As a remedy, we propose to adopt an active TL paradigm, where we purposely select the most valuable cancer messages and query the new GURI with these valuable messages instead of all unlabelled messages from the candidate dataset. Active selection minimises the number of queries on the target GURI while maintaining high-quality fine-tuning. EvoCLINICAL builds an effective CCDT (denoted as CCDT-T) for the target GURI with the knowledge transferred from the source GURI's CCDT (denoted as CCDT-S). Figure 1 shows EvoCLINICAL's overview comprising two stages: *CCDT-S Construction Stage* and *CCDT-T Construction Stage*. *CCDT-S Construction Stage* trains CCDT-S for GURI-S to predict rule validation results with the dataset collected during the operation of GURI-S. *CCDT-T Construction Stage* entails *Evolution Dataset Construction Phase* and *Transfer Learning Phase*. During the first phase, we construct a dataset for evolution by selecting a subset of valuable cancer messages from the candidate dataset. Specifically, we define five objectives and employ a genetic search algorithm (i.e., IBEA) to find an optimal subset of cancer messages. Next, we query GURI-T to validate each message in the subset, which is then up-sampled into a larger dataset $CM$ to increase the training data volume. Finally, in *Transfer Learning Phase*, we build the CCDT-T by extending CCDT-S by adding new model modules dedicated to newly-introduced rules. We represent the parameters for previously existing and new modules as $P_{CCDT-S}$ and $P_{New}$, respectively. We consider CCDT-S as the pretrained model for CCDT-T, sharing its trained parameters as $P_{CCDT-S}$'s initialization. We randomly initialise new modules' parameters since CCDT-S does not contain knowledge about $P_{New}$. After initialization, we use dataset $CM$ to fine-tune $P_{CCDT-S}$ and train $P_{New}$ from scratch.

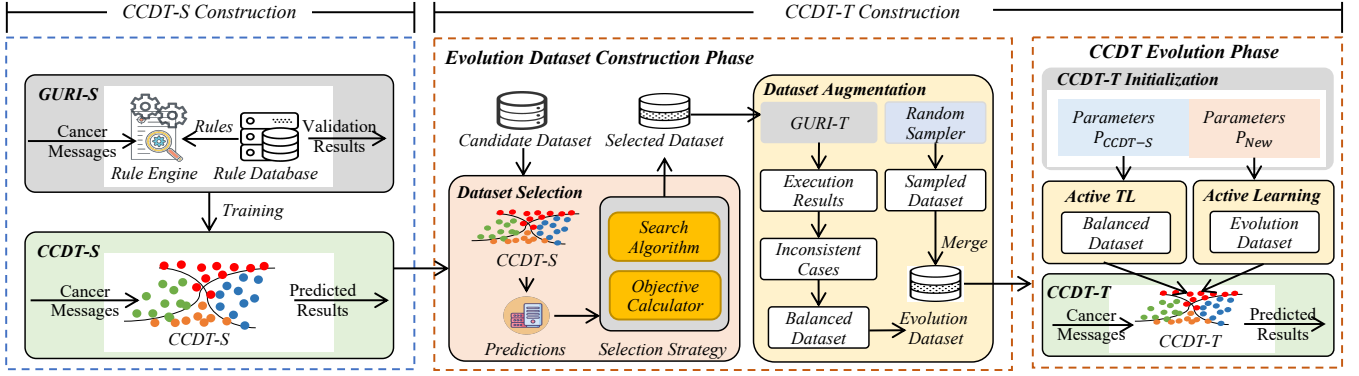

Figure 1: Overview of EvoCLINICAL. GURI-S denotes the source GURI version, which evolves to GURI-T, the target GURI version. CCDT-S and CCDT-T are the cyber-cyber digital twins of GURI-S and GURI-T, respectively. TL denotes transfer learning.

### 3.1 Stage 1: CCDT-S Construction

As mentioned in Section 2, GURI validates a given cancer message $X$ with rule $i$, labelling it as $y^i \in$ {"info", "warning", "not applied", "error"}. To simulate GURI-S, we design CCDT-S as a 4-category classification neural network model. Specifically, we employ a multi-output classification model [38] as the CCDT-S to predict the validation result of a cancer message. Figure 2 shows the detailed structure of CCDT-S. For a given cancer message $X$, we first apply a series of feature preprocessing techniques to preprocess the inputs as vector representations, which will be inputted to CCDT-S. CCDT-S is a multi-output model consisting of several independent prediction modules with identical architecture, where each module is responsible for the result code prediction of one specific rule. Let the rule set size of GURI-S be $n$. We design CCDT-S with $n$ modules, i.e., $M_1, M_2, ..., M_n$. Next, we present the design of CCDT-S: its feature representation (Section 3.1.1) and architecture (Section 3.1.2).

#### 3.1.1 Feature Representation. 
A cancer message $X$ encompasses multiple variables of the same or different types, i.e., numerical, categorical, and string types. Therefore, feature representation techniques are needed to transform the original cancer message dataset into numerical feature representations, such that they can be processed by CCDT-S. Specifically, we apply the following feature representation approaches for handling different variable types.

**One-hot encoding** is a frequently used approach to deal with categorical data in machine learning [37]. In our context, for variables with $nc$ categorical values, we represent each categorical value as a $nc$-dimensional sparse vector where all elements are 0 except for a single 1 that uniquely identifies the categorical value.

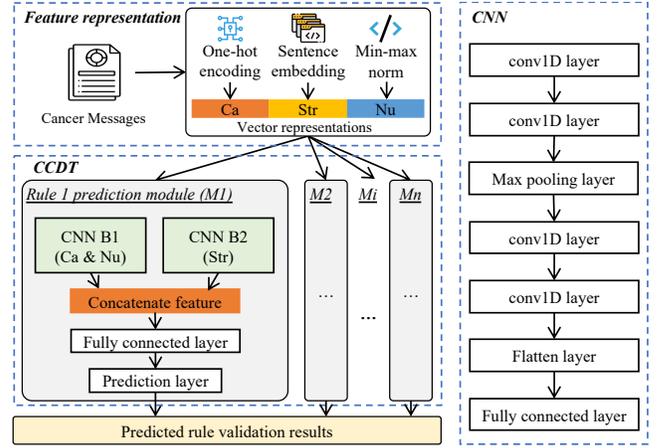

Figure 2: Architecture of CCDT. Ca: categorical variables; Str: string variable; Nu: numerical variables

**Sentence embedding** is a natural language processing (NLP) technique, which represents entire sentences and their semantic information as vectors [36]. It is trained on various data sources. It can encode sentences, phrases, or short paragraphs and is commonly used to identify similar texts based on context, meaning, and topics. We choose universal sentence encoder (USE) [10], a commonly applied sentence embedding model, to represent each string variable value as a 512-dimensional vector.

**Min-max normalization**, also known as min-max scaling, is applied to scale numerical variable values within the range of [0, 1]. This can ensure that numerical variables contribute equally to the model fitting as categorical and string variables.

#### 3.1.2 Module Architecture. 
As we discussed earlier, for a GURI version with $n$ rules, its corresponding CCDT will have a CCDT-S with $n$ modules, each of which is responsible for one specific rule result prediction. In our current design of CCDT-S, all the modules have the same module architecture, and we employ the multi-input CNN to design a module, which has been proven effective for breast cancer detection [40]. Specifically, as shown in Figure 2, a module has two CNN branches, where CNN B1 is for categorical and numerical variable values, and CNN B2 is for string variable values. Each branch processes its input (vector representations) independently. Outputs of the two branches are concatenated and processed by a fully connected layer. Its output is then processed by the prediction layer, which uses softmax as the activation function to obtain the prediction result.

CNN B1 and CNN B2 share the same architecture shown on the right-hand side of Figure 2. The first two layers are one-dimensional convolutional (conv1D) layers with the kernel size being 16 and 32, respectively. The third layer is the max pooling layer with stride 2, followed by two conv1D layers of kernel size 64. The flattening layer is used to flatten the output of the previous layer. The last layer is a fully connected layer with 200 neurons.

## 3.2 Stage 2: CCDT-T Construction

To construct CCDT-T, we first utilize search-based selection to construct a subset $CM$ of valuable cancer messages (Section 3.2.1). Then we harness TL to transfer knowledge from CCDT-S to CCDT-T by fine-tuning with dataset $CM$ (Section 3.2.2).

### 3.2.1 Evolution Dataset Construction Phase.
This phase involves dataset selection and augmentation.

**Dataset Selection** is formulated as a search problem. Let $CM_{S \to T} = \{X_1, X_2, ..., X_{nc}\}$ be a cancer message dataset, where $nc$ is the total number of cancer messages in $CM_{S \to T}$. A cancer message $X$ is specified with several variables related to cancer registration: $X = \{var_1, var_2, ..., var_{nv}\}$. Notice that $CM_{S \to T}$ can be obtained by a cancer message generation strategy or provided by CRN, and the generation strategy is out of the scope of this paper. Our objective is to search a subset of cancer messages from $CM_{S \to T}$ that satisfies a specific optimization goal to evolve CCDT from CCDT-S to CCDT-T. Therefore, the entire search space is all non-empty subsets of $CM_{S \to T}$, and we define the search space of all possible solutions as: $PS_{S \to T} = \{ps_1, ps_2, ..., ps_{nps}\}$, where $nps$ is the total number of possible solutions. Notice that $nps$ can be calculated as $2^{nc} - 1$, and as the size of $CM_{S \to T}$ (i.e., $nc$) increases, the search space will exponentially increase and exhaustively explore the entire search space is practically infeasible.

Each possible solution $ps_i \in PS_{S \to T}$ has $nps_i$ ($1 \le nps_i \le nc$) cancer messages: $ps_i = \{X_1, X_2, ..., X_{nps_i}\}$. For a cancer message $X_j \in ps_i$, we can use CCDT-S to predict the validation results of each rule in $R_S$, which can be represented as a set of probability vectors: $PV_{X_j} = \{pv_{X_j}^{r_1}, pv_{X_j}^{r_2}, ..., pv_{X_j}^{r_{nr_S}}\}$, where $nr_S$ is the size of $R_S$. Each probability vector has four probability values denoting the probabilities of the four types of result codes (i.e., "info", "warning", "not applied", "error"). Therefore, for solution $ps_i$, CCDT-S predicts its validation results as: $PVR_{ps_i} = \{PV_{X_1}, PV_{X_2}, ..., PV_{X_{nps_i}}\}$. Notice that the sum of the four probability values in a probability vector is one, and we take the position with the maximum probability value as the predicted rule result code, i.e., arg max. For example, $pv_{X_j}^{r_1} = [0.1, 0.05, 0.65, 0.3]$, then the predicted result code for rule $r_1$ is 2, which is "not applied".

Furthermore, we can also get the true validation results of solution $ps_i$ on GURI-S: $TVR_{ps_i}^{GURI_S} = \{tvr_{X_1}, tvr_{X_2}, ..., tvr_{X_{nps_i}}\}$, where $tvr_{X_j}$ is the true validation results of the $j_{th}$ cancer message (i.e., $X_j$) in $ps_i$.

**Solution Measurements.** To measure the goodness of a solution $ps$ in $PS_{S \to T}$, we define the following set of solution measurements.

1) *Solution Size (SS)* measures the amount of selected cancer messages in solution $ps$. SS is calculated as:

$$SS_{ps} = nps, \quad (1)$$

where $nps$ is the total number of cancer messages in solution $ps$.

2) *Cancer Message Diversity (CMD)* measures the diversity of solution $ps$. To calculate $CMD$, we first define the difference of two cancer messages $X_j, X_k$ in $ps$:

$$DIV_{j,k} = \frac{count_{identical}(X_j, X_k)}{nv}, \quad (2)$$

where, function $count_{identical}$ counts the number of variables having the same values in $X_j$ and $X_k$, and $nv$ is the total number of variables in cancer messages. Then we define $CMD$ of solution $ps$ as the average of the difference of each pair of messages in $ps$:

$$CMD_{ps} = \frac{\sum_{j=1}^{nps-1} \sum_{k=j+1}^{nps} DIV_{j,k}}{\sum_{m=1}^{nps-1} m}, \quad (3)$$

where $nps$ is the total number of cancer messages in $ps$.

3) *Result Code Diversity (RCD)* measures the similarity between the four result code distributions between $ps_i$ and the original dataset $CM_{S \to T}$, with the Jensen–Shannon (JS) divergence [17], a method of measuring the similarity between two probability distributions, $P$ and $Q$:

$$JS(P||Q) = \frac{1}{2} * KL(P||M) + \frac{1}{2} * KL(Q||M), \quad (4)$$

where $M$ is a probability distribution calculated as $1/2 * (P + Q)$, and $KL(P||M)$ is the Kullback-Leibler (KL) divergence between distributions $P$ and $M$:

$$KL(P||M) = \sum_{x \in X} P(x) log(\frac{M(x)}{P(x)}). \quad (5)$$

Based on JS-divergence, we calculate $RCD$ between the result code distributions achieved by $ps$ and $CM_{S \to T}$:

$$RCD_{ps} = -1 * \frac{\sum_{j=1}^{nr_s} JS(D_{r_j}^{ps} || D_{r_j}^{CM_{S \to T}})}{nr_s}, \quad (6)$$

where, $nr_S$ is the number of validation rules in GURI-S, and $D_{r_j}^{ps}$ and $D_{r_j}^{CM_{S \to T}}$ are the result code distributions of rule $r_j$ in $ps$ and $CM_{S \to T}$, respectively.

4) *False Prediction Proportion (FPP)* is to measure how many cancer messages in $ps$ lead to predictions by CCDT-S to differ from the real validation results in GURI-S:

$$FPP_{ps} = \frac{\#FP}{nps}, \quad (7)$$

where $\#FP$ is the number of cancer messages that lead to predictions to differ from real validation results in GURI-S.

5) *Prediction Uncertainty (PU)* measures how much confidence CCDT-S have on its predictions. To calculate $PU$, we employ information entropy [6] as the uncertainty quantification measures:

$$H(X) = -\sum_{x \in X} P(x) \log P(x) = \mathbb{E}[-\log P(X)], \quad (8)$$

where $X$ is a set of events, $P$ is the probability distribution of $X$, and $P(x)$ is the probability of event $x \in X$.

Information entropy calculates the amount of information in $X$. A smaller value of information entropy implies a higher uncertainty in $X$. Based on information entropy, we first calculate the prediction uncertainty of one cancer message $X_i \in ps$ as:

$$CMPU_i = \frac{\sum_{j=1}^{nr_S} H(pv_{X_i}^{r_j})}{nr_S}, \quad (9)$$

where $pv_{X_i}^{r_j}$ is the probability vector of rule $r_j$ of the prediction of $X_i$, and $nr_S$ is the total number of rules applied in GURI$_S$. Then, for solution $ps$, we calculate its $PU$ as:

$$PU_{ps} = \frac{\sum_{i=1}^{nps} CMPU_i}{nps}. \quad (10)$$

**Problem Formulation.** Since we aim to evolve CCDT-S to CCDT-T by applying active TL, the quality of the dataset for evolving CCDT-S is the key to obtaining a high quality CCDT-T with the least possible cost. Therefore, we want to select a dataset, as a possible solution $ps$, with the minimum size. Our optimization problem can be represented as follows: Given a set of cancer messages $CM_{S \to T}$ with $nc$ being its size, find solution $ps_k \in PS_{S \to T}$ that satisfies: $\forall ps_i \in PS_{S \to T} \cap ps_i \neq ps_k$:

(1) $SS_{ps_k} \leq SS_{ps_i}$, implying that $ps_k$ has the minimum number of cancer messages. This is important because we want to obtain CCDT-T with minimum cost.

(2) $CMD_{ps_k} \geq CMD_{ps_i}$, indicating that cancer messages in $ps_k$ have the most diversity. A higher diversity means lower duplication of cancer messages in the dataset.

(3) $RCD_{ps_k} \geq RCD_{ps_i}$, showing that the prediction results of $ps_k$ have the most diverse distributions. This objective is to ensure that the predictions are diverse and close to the distribution from the candidate dataset.

(4) $FPP_{ps_k} \geq FPP_{ps_i}$, showing that in $ps_k$, the proportion of cancer messages with false prediction is the highest. Recall that we want to obtain high-quality CCDT-T by evolving CCDT-S. Therefore, cancer messages that cause false predictions of CCDT-S are expected to be more important for evolving CCDT-S and obtaining a high performance CCDT-T.

(5) $PU_{ps_k} \geq PU_{ps_i}$, implying that the prediction of cancer messages in $ps_k$ has the highest uncertainty. Notice that the cancer messages in the dataset are not all valuable in terms of evolving a CCDT from CCDT-S to CCDT-T, and according to [42], data with higher uncertainty is more useful in improving the prediction performance of a model. This inspires us to select the most uncertain samples for TL.

**Dataset Augmentation.** As Figure 1 shows, with the selected dataset, we augment it to get the final dataset to evolve CCDT-S. Concretely, we first execute the selected dataset on GURI-T and identify cancer messages that have CCDT-S's prediction results different from the execution results. Then, we balance the selected dataset by upsampling or downsampling these inconsistent cancer messages to get a balanced dataset. In addition, considering that GURI-T has newly added modules, we need to train the parameters of these modules from scratch. To do so, we further randomly sample a dataset from cancer messages that are pre-generated or provided by CRN and merge it with the balanced dataset. Finally, we obtain an *evolution dataset*, which includes the *balanced dataset* and *sampled dataset*. The *evolution dataset* will be used in *Evolution Phase* for TL.

*3.2.2 Transfer Learning Phase.* We denote the rule set for GURI-S as the base rules $R_B$ and new rules $R_E$. Correspondingly, CCDT-T consists of base modules $P_{CCDT-S}$ and new modules $P_{New}$, each of which predicts the validation result of each rule in GURI-T. The last box of Figure 1 illustrates the training process of the base and new modules. For the base modules, we employ active TL for their training, fine-tuning the pretrained model with actively selected cancer messages with labels from GURI-T. For the new modules, we adopt the same actively selected cancer messages to train $P_{New}$ from scratch.

## 4 EXPERIMENT DESIGN

We present research questions (RQs) (Section 4.1), experiment settings (Section 4.2), and evaluation metrics (Section 4.3).

### 4.1 Research Questions

Recall that EvoCLINICAL evolves a source CCDT to a target CCDT by fine-tuning the pretrained source CCDT with labelled data from the target GURI. Hence, to evaluate EvoCLINICAL, we pose three RQs: **RQ1**: Is EvoCLINICAL effective in evolving CCDT of GURI? **RQ2**: How effective is the active learning paradigm in contributing to the performance of EvoCLINICAL? **RQ3**: To what extent does the candidate dataset size affect the performance of EvoCLINICAL?

TL is used for evolving source CCDT to the target CCDT. In *RQ1*, we consider two baselines without TL for target CCDT construction. The first baseline utilizes the pretrained source CCDT as the target CCDT without fine-tuning, referred to as OTS. The second baseline trains a target CCDT from scratch, denoted as TFS. *RQ2* investigates active learning's contribution to EvoCLINICAL's performance. Active learning selects a subset of valuable cancer messages to query GURI-T and train CCDT-T. Alternatively, such a subset can be selected randomly without considering the importance of each cancer message. To demonstrate active learning's effectiveness, we compare training CCDT-T with randomly selected (denoted as EvoCLINICAL-RS) and actively selected cancer messages (i.e., EvoCLINICAL). Since active learning is adopted for the training of base modules $P_{CCDT-S}$ and new modules $P_{New}$, we evaluate its effectiveness separately on $P_{CCDT-S}$ and $P_{New}$. *RQ3* studies the influence of candidate dataset size on EvoCLINICAL's performance.

### 4.2 Experiment Settings

We determine the hyperparameters of CCDT by experimenting with different combinations on a reserved validation dataset. As a result, we use Adam [23] as the optimizer with a learning rate of 0.001. We build the neural network models with Tensorflow Framework 2.11.0. In terms of the multi-objectives optimization algorithm, we employ the indicator-based evolutionary algorithm IBEA [45] because evidence has shown that it is effective in solving multi-objective problems [29]. We also used the default parameter settings of IBEA from jMetal [13] except for the number of evaluations set to 30000 based on our pilot study. All experiments are conducted on one server node equipped with an Intel Xeon Platinum 8186, 16x NVIDIA V100 GPU.

### 4.3 Evaluation Metrics and Statistical Testing

We adopt three commonly used classification metrics: precision, recall, and F1 score, which are defined based on the number of true positives, false positives, false negatives, and false positives denoted as $TP$, $FP$, $FN$, and $FP$, respectively. Precision measures how accurate the model is for positive predictions, as calculated in Equation 11.

$$precision = \frac{TP}{TP + FP} \quad (11)$$

As shown in Equation 12, recall assesses how well the model identifies true positives.

$$recall = \frac{TP}{TP + FN} \quad (12)$$

Equation 13 shows that the F1 score is computed as the harmonic mean of precision and recall, taking both false positives and false negatives into account.

$$F_1 = 2 \cdot \frac{precision \cdot recall}{precision + recall} \quad (13)$$

To reduce the effect of randomness on the experiment's validity, we repeat each experiment 10 times and perform statistical testing to demonstrate the significance of each difference. Specifically, we adopt the Mann-Whitney test as suggested in [2]. Furthermore, we calculate Vargha-Delaney's $\hat{A}_{12}$ to estimate the effect size of each difference. $\hat{A}_{12}$ ranges from 0 to 1 and requires no knowledge of the data distribution. When comparing method A with method B, a higher $\hat{A}_{12}$ value (> 0.5) indicates method A has a higher chance of yielding better results and vice versa.

## 5 EXPERIMENT RESULTS

In this section, we present the results for each research question. The replication package is available in our GitHub repository[1].

### 5.1 RQ1 - Overall Effectiveness of EvoCLINICAL

Table 2 demonstrates the effectiveness of EvoCLINICAL in three evolution processes, namely $S_1 \to T_1$, $S_2 \to T_2$, and $S_3 \to T_3$. As presented in the last row, the F1 scores of EvoCLINICAL on the three evolution processes are 0.9338, 0.9137 and 0.9233, respectively. This result indicates that EvoCLINICAL is comprehensively effective in constructing target CCDT-T. The precision values on the three evolution processes are all greater than 0.9200, which means more than 92% result code predictions are correct. Furthermore, the lowest recall value recorded is 0.9105, indicating that EvoCLINICAL manages to predict at least 91.05% of the result codes successfully.

Essentially, EvoCLINICAL uses CCDT-S as a pretrained model and fine-tunes it with actively selected cancer messages to construct CCDT-T with TL. We dissect the TL approach and study the individual contribution of pretraining and fine-tuning to the overall effectiveness of EvoCLINICAL. We compare EvoCLINICAL with OTS, which trains a CCDT-T from scratch without relying on the pretrained CCDT-S (Section 4.1). Results are presented in Row 2 of Table 2. We find all the metrics plummet from above 0.9 to below 0.8, with a maximum decrease of 0.1966 on the recall value of $S_3 \to T_3$. Table 3 demonstrates the significance (p − value < 0.05) of such decreases with large effect sizes ($\hat{A}_{12} = 1$). These results align with our expectations since the pretrained CCDT-S possesses valuable information for the construction of CCDT-T. Specifically, we leverage this information for a more optimal initialization of CCDT-T. By contrast, OTS randomly initializes CCDT-T's parameters without prior knowledge, which leads to significant decreases in the performance. Considerably more cancer messages and training iterations are required in the future to fill the knowledge gap.

Meanwhile, we analyze the effectiveness of fine-tuning by comparing EvoCLINICAL with OTS, where CCDT-S is utilized directly without fine-tuning. Table 2 shows a significant decrease in all three evolution processes regarding precision, recall, and F1 score, with all the p-values less than 0.05 and all the $\hat{A}_{12}$ values being 1.0. This is reasonable since the fine-tuning trains the pretrained model with cancer messages labelled by GURI-T. These labelled cancer messages contain unique information of GURI-T, which was, however, not available when CCDT-S was trained. Training with these labelled messages incorporates GURI-T's information into the CCDT-T model. Consequently, the fine-tuned CCDT-T model can produce better results compared to OTS.

### 5.2 RQ2 - Effectiveness of Active Learning Paradigm

Recall that active learning selects cancer messages to fine-tune the base modules, each of which corresponds to a rule in CCDT-S. To test the contribution of active learning in the overall performance of EvoCLINICAL, we also present the performance of fine-tuning with the random selection strategy (i.e., EvoCLINICAL-RS) in Table 2 as the comparison baseline. Results show that EvoCLINICAL outperforms EvoCLINICAL-RS in terms of all metrics in all three evolution processes. Table 3 further shows the significance of all the increments, except for precision in $S_1 \to T_1$ and $S_2 \to T_2$. Despite the insignificance, we still find small effect sizes in terms of precision: $\hat{A}_{12} = 0.64$ in $S_1 \to T_1$ and $\hat{A}_{12} = 0.65$ in $S_2 \to T_2$ and. In short, EvoCLINICAL still has a better chance of yielding better results in precision in $S_1 \to T_1$ and $S_2 \to T_2$, compared to EvoCLINICAL-RS.

Recall that active learning is utilized for training new modules from scratch. Table 4 compares training new modules (corresponding to new rules introduced to CCDT-T) with EvoCLINICAL-RS and actively selected cancer messages (EvoCLINICAL). EvoCLINICAL's performance for all the new rules outperforms EvoCLINICAL-RS regarding precision, recall, and F1 score. The largest increase reaches 0.0366 on Recall for $S_2 \to T_2$ for $R1_{T_2}$. Furthermore, we find the overall performance displayed in Table 4 decreases by a large margin in terms of all three metrics, compared to the performance on the base modules from Table 2. This decrease lives up to our expectations since the training on the base modules can leverage the pretrained CCDT-S for knowledge transfer, which is not the case for the new modules. In a word, active learning is effective in fine-tuning base modules and training new modules from scratch.

---
[1]https://github.com/Simula-COMPLEX/EvoCLINICAL

| Method | $S_1 \to T_1$ | | | $S_2 \to T_2$ | | | $S_3 \to T_3$ | | |
|---|---|---|---|---|---|---|---|---|---|
| | *Precision* | *Recall* | $F_1$ | *Precision* | *Recall* | $F_1$ | *Precision* | *Recall* | $F_1$ |
| TFS | 0.7722▲ | 0.7426▲ | 0.7518▲ | 0.7434▲ | 0.7095▲ | 0.7187▲ | 0.7566▲ | 0.725▲ | 0.7324▲ |
| OTS | 0.9272 ▲ | 0.9237 ▲ | 0.9244▲ | 0.9125▲ | 0.8981▲ | 0.9035▲ | 0.9106▲ | 0.9146▲ | 0.9116▲ |
| EvoCLINICAL-RS | 0.9376 | 0.9287▲ | 0.9313▲ | 0.9185 | 0.9069▲ | 0.9105▲ | 0.9247▲ | 0.9201▲ | 0.9207▲ |
| **EvoCLINICAL** | **0.9389** | **0.9313** | **0.9338** | **0.9200** | **0.9105** | **0.9137** | **0.9281** | **0.9216** | **0.9233** |

Table 2: Results of comparing EvoCLINICAL and the baselines on the base rules in terms of precision, recall and F1 score. OTS denotes using the pretrained source CCDT off-the-shelf; TFS denotes training a target CCDT from scratch; EvoCLINICAL-RS represents using a random search strategy instead of active learning to select the fine-tuning dataset; ▲ denotes that EvoCLINICAL is significantly better than a baseline model, i.e., $\hat{A}_{12} > 0.5, p-value < 0.05$.

| Evolution Pair | Baseline | *Precision* | | *Recall* | | $F_1$ | |
|---|---|---|---|---|---|---|---|
| | | $\hat{A}_{12}$ | $p$ | $\hat{A}_{12}$ | $p$ | $\hat{A}_{12}$ | $p$ |
| $S_1 \to T_1$ | TFS | **1.0** | **<0.05** | **1.0** | **<0.05** | **1.0** | **<0.05** |
| | OTS | **1.0** | **<0.05** | **1.0** | **<0.05** | **1.0** | **<0.05** |
| | EvoCLINICAL-RS | 0.64 | 0.31 | **0.87** | **<0.05** | **0.81** | **<0.05** |
| $S_2 \to T_2$ | TFS | **1.0** | **<0.05** | **1.0** | **<0.05** | **1.0** | **<0.05** |
| | OTS | **1.0** | **<0.05** | **1.0** | **<0.05** | **1.0** | **<0.05** |
| | EvoCLINICAL-RS | 0.65 | 0.27 | **0.9** | **<0.05** | **0.93** | **<0.05** |
| $S_3 \to T_3$ | TFS | **1.0** | **<0.05** | **1.0** | **<0.05** | **1.0** | **<0.05** |
| | OTS | **1.0** | **<0.05** | **1.0** | **<0.05** | **1.0** | **<0.05** |
| | EvoCLINICAL-RS | **0.84** | **<0.05** | **0.76** | **<0.05** | **0.94** | **<0.05** |

Table 3: Results of the statistical tests comparing EvoCLINICAL and the baselines on the base rules. OTS denotes using the pretrained source CCDT off-the-shelf; TFS denotes training a target CCDT from scratch; EvoCLINICAL-RS represents using a random search strategy instead of active learning to select the fine-tuning dataset.

## 5.3 RQ3 - Impact of Candidate Dataset Size on EvoCLINICAL's Effectiveness

Table 5 reports the experiment results of EvoCLINICAL under different candidate dataset sizes, ranging from 1000 to 8000. We find the performance of EvoCLINICAL tends to increase as the candidate dataset size grows. Comparing sizes 1000 and 8000, the F1 scores increase 0.064 (0.7160-0.6554), 0.0484 (0.8527-0.8043), and 0.0847 (0.8273-0.7426) in $S_1 \to T_1$, $S_2 \to T_2$, and $S_3 \to T_3$, respectively.

Another observation from the table is that there is a progressive deceleration in the growth rate. Such deceleration can be derived from two aspects: data quality and model capacity. The dataset quality might decrease as newly added cancer messages to the candidate datasets can be redundant or less relevant to the validation task. Another reason for the deceleration lies in the model capacity. A neural network with a small capacity (i.e., EvoCLINICAL) tends to overfit on a large dataset, decreasing the model performance.

The deceleration implies the diminishing returns of keeping increasing the candidate sizes. A suitable candidate dataset size should be determined empirically or based on prior knowledge to reduce the query times on GURI-T.

## 6 DISCUSSION AND LESSONS LEARNED

*Benefiting from transfer learning.* An effective TL strategy paves the way for automatic CCDT evolution. Instead of training a CCDT from scratch, TL leverages the source CCDT and transfers knowledge to the target CCDT. In this paper, TL improves the performance of EvoCLINICAL compared to training from scratch. The plausible reasons for such improvement are discussed as follows. (1) TL transfers mutual knowledge from the source CCDT to the target DT. In the context of GURI, this mutual knowledge mainly entails the knowledge of feature extraction. Despite the update of GURI, the structure of a cancer message remains the same. Therefore, the feature extraction capability of the source CCDT can be harnessed to improve that of the target CCDT. (2) From an optimization perspective, the source CCDT provides a more stable and reasonable location for the parameter initialization of the target CCDT. As Sutskever et al. [39] pointed out, the initialization of deep learning models is crucial for subsequent optimization. (3) TL can also be considered a form of regularization, incorporating an inductive bias derived from the source CCDT into the target CCDT. This inductive bias can help reduce the risk of overfitting and improve the generalizability of target CCDT.

*Enabling testing.* Testing of GURI to ensure its dependability is essential. Several automated testing practices have been performed on GURI. Specifically, Laaber et al. [24] utilized an AI-based system-level testing tool, EvoMaster [3] to test GURI. Isaku et al. [20] proposed to use a machine learning classifier to filter out test requests that won't lead to rule executions, consequently reducing the cost of testing GURI. Despite the promising results of these works, Laaber et al. [25] pointed out that it might be beneficial to test GURI by learning and simulating medical coders. A CCDT can enable such testing. Ahlgren et al. [1] stated that a well-established CCDT could be considered a true twin of the system of concern, informing and affecting each other. Such a bidirectional connection enables many applications, such as regression testing, run-time verification, and preventing real-time system failures. For example, given an unlabeled cancer message $X$, CCDT and GURI validate $X$ simultaneously. If the validation results are at odds and no test oracles are available, we consider the risk of the system failing to process the cancer message high, and developers are therefore informed to perform manual examinations. Thus, EvoCLINICAL can be used to select cancer messages (i.e., test inputs) with a high probability of failing GURI.

## 7 RELATED WORK

We discuss the related works covering these three topics: cyber-cyber digital twin (Section 7.1), transfer learning (Section 7.2), and active learning (Section 7.3).

| Method | | $S_1 \to T_1$ | | | | $S_2 \to T_2$ | | | | $S_3 \to T_3$ | | |
| --- | --- | --- | --- | --- | --- | --- | --- | --- | --- | --- | --- | --- |
| | | Precision | Recall | $F_1$ | | Precision | Recall | $F_1$ | | Precision | Recall | $F_1$ |
| EvoCLINICAL-RS | $R1_{T_1}$ | 0.8054 | 0.7020 | 0.7265 | $R1_{T_2}$ | 0.7825 | 0.7225 | 0.7345 | $R1_{T_3}$ | 0.9267 | 0.9396 | 0.9329 |
| EvoCLINICAL | | 0.8345 | 0.7164 | 0.7454 | | 0.7963 | 0.7591 | 0.7696 | | 0.9304 | 0.9431 | 0.9366 |
| EvoCLINICAL-RS | $R2_{T_1}$ | 0.7316 | 0.7225 | 0.7255 | $R2_{T_2}$ | 0.7969 | 0.7674 | 0.7795 | $R2_{T_3}$ | 0.8591 | 0.8351 | 0.8420 |
| EvoCLINICAL | | 0.7321 | 0.7251 | 0.7279 | | 0.8082 | 0.7794 | 0.7904 | | 0.8617 | 0.8280 | 0.8361 |
| EvoCLINICAL-RS | $R3_{T_1}$ | 0.5912 | 0.4979 | 0.5323 | $R3_{T_2}$ | 0.8116 | 0.8244 | 0.8166 | $R3_{T_3}$ | 0.6880 | 0.6062 | 0.6348 |
| EvoCLINICAL | | 0.6014 | 0.5012 | 0.5375 | | 0.8199 | 0.8413 | 0.8298 | | 0.6837 | 0.5958 | 0.6258 |
| EvoCLINICAL-RS | $R4_{T_1}$ | 0.7868 | 0.7883 | 0.7869 | $R4_{T_2}$ | 0.9460 | 0.9802 | 0.9625 | $R4_{T_3}$ | 0.8792 | 0.8459 | 0.8604 |
| EvoCLINICAL | | 0.7949 | 0.7884 | 0.7889 | | 0.9496 | 0.9828 | 0.9657 | | 0.8867 | 0.8533 | 0.8674 |
| EvoCLINICAL-RS | $R5_{T_1}$ | 0.7894 | 0.7485 | 0.7650 | $R5_{T_2}$ | 0.9481 | 0.9345 | 0.9409 | $R5_{T_3}$ | 0.8898 | 0.8059 | 0.8377 |
| EvoCLINICAL | | 0.8102 | 0.7613 | 0.7805 | | 0.9431 | 0.9292 | 0.9357 | | 0.9263 | 0.8356 | 0.8704 |

Table 4: Results of comparing EvoCLINICAL and the baselines on the new rules regarding precision, recall, and F1 score. EvoCLINICAL-RS uses a random search strategy instead of active learning to select the fine-tuning dataset; $Ri_{T_j}$ is the $ith$ newly introduced rule in target GURI-$T_j$.

| Candidate Dataset Size | $S_1 \to T_1$ | | | $S_2 \to T_2$ | | | $S_3 \to T_3$ | | |
| --- | --- | --- | --- | --- | --- | --- | --- | --- | --- |
| | Precision | Recall | $F_1$ | Precision | Recall | $F_1$ | Precision | Recall | $F_1$ |
| 1000 | 0.6827 | 0.6423 | 0.6554 | 0.8237 | 0.7984 | 0.8043 | 0.7873 | 0.7218 | 0.7426 |
| 2000 | 0.6907 | 0.6403 | 0.6562 | 0.8225 | 0.8034 | 0.8087 | 0.7997 | 0.7434 | 0.7616 |
| 3000 | 0.7039 | 0.6558 | 0.6691 | 0.8461 | 0.814 | 0.8204 | 0.8102 | 0.7526 | 0.7722 |
| 4000 | 0.7173 | 0.6716 | 0.6861 | 0.8389 | 0.8263 | 0.8276 | 0.8217 | 0.7656 | 0.7844 |
| 5000 | 0.7376 | 0.6798 | 0.6971 | 0.8442 | 0.8228 | 0.8279 | 0.8312 | 0.7866 | 0.8022 |
| 6000 | 0.7465 | 0.6923 | 0.7085 | 0.8529 | 0.8396 | 0.8415 | 0.8431 | 0.795 | 0.8119 |
| 7000 | 0.7448 | 0.6932 | 0.7095 | **0.8654** | 0.8482 | 0.8514 | 0.8534 | **0.8166** | **0.8292** |
| 8000 | **0.7546** | **0.6985** | **0.7160** | 0.8577 | **0.8533** | **0.8527** | **0.8578** | 0.8112 | 0.8273 |

Table 5: Results of EvoCLINICAL's performance under different candidate dataset sizes regarding precision, recall, and F1 score.

## 7.1 Cyber-cyber Digital Twin

The concept of DT has been intensively studied in both academia and industry in recent years. Various DTs have been designed and deployed for different contexts, such as health monitoring [44], product lifecycle management, and cyber-physical systems security. El Saddik [14] defined a concept of *Digital Twin* as *"a digital replica of a living or non-living physical entity"*. This definition bound the concept of DT to physical entities, while complex software systems can also benefit from it.

Facebook proposed to build a CCDT named WW for their WWW platform [1]. WW simulates the user interaction and their platform. The authors demonstrated three key advantages of CCDT, namely complete malleability, true twins, and simulation hierarchy [1]. Complete malleability means that, theoretically, no change is unimplementable in a software system, which is not valid in a physical system. Moreover, CCDT and its corresponding software systems can become true twins since they either inform or affect the other, thanks to their malleability. Simulation hierarchy indicates the possibility of building a CCDT for a CCDT since a CCDT can also be considered a software system. GURI, the system under study in this paper, can also capitalize on these advantages. Therefore, we follow this research line and build CCDT for GURI.

## 7.2 Transfer learning

TL entails applying knowledge gained from solving one task to another related task. Most recent developments of TL originated in the 2010s when the pretraining+fine-tuning paradigm was proposed to classify ImageNet pictures. The pretrained model improves the performance of image classification and the downstream tasks in the computer vision domain, such as image segmentation [41] and object detection [19].

The vast success of ImageNet illuminates a myriad of research on applying TL in other domains, especially natural language processing [11]. In particular, language models (LMs) are increasingly studied as a counterpart for ImageNet in the natural language processing domain. Pretrained on a large corpus, an LM can produce meaningful embeddings to facilitate downstream tasks, such as sentiment analysis [26], question answering [32], and named entity recognition [28]. Mikolov et al. proposed to use shallow neural networks to generate embeddings for each token [31], followed by a series of other embedding techniques, e.g., fasttext [22] and glove [33]. Later on, researchers dedicated their efforts to incorporating context information with the pretrained language models by proposing contextualized language models such as ElMo [34], BERT [12], and GPT [7]. The phenomenal chat bot ChatGPT is also constructed based on the GPT language model.

Despite its wide applications in natural language processing and computer vision, TL has been utilized by a few researchers to evolve a digital twin. One plausible reason is the relatively low data availability in the software/physical systems compared to computer vision and natural language processing domains. Xu et al. [42] built RISE-DT for an elevator system and utilized TL to evolve it for different traffics and versions. RISE-DT tackles time-series data consisting of sensor and actuator values, which differs significantly from the cancer messages. Xu et al. [43] also proposed a DT-based method, named KDDT, for anomaly detection in the network of a train control and management system. KDDT enriched the training data for DT by distilling knowledge from the out-of-domain dataset. In this paper, we design an entirely different architecture of EvoCLINICAL to evolve CCDT for different GURI versions and employ active learning to improve the TL efficiency further.

### 7.3 Active Learning

Active learning is particularly useful when the unlabeled data samples are abundant, whereas manually labelling all of them is expensive or infeasible. Active learning is widely used in many areas, such as text classification [18], information extraction [9], image classification [21], and speech recognition [30].

Cai et al. [8] proposed to select the most informative samples based on the prediction uncertainty. They demonstrate the effectiveness of the proposed method in the text classification task. However, their method suffers from sampling bias since the distribution of the selected instances can differ from the original distribution. To mitigate this problem, several approaches take the diversity of selected data into consideration [27, 35]. The basic idea is to avoid selecting similar instances for annotation that increase the labelling cost. Another issue active learning faces is failing to select minority classes in an imbalanced dataset, especially in a multi-output task [4]. This paper proposes tackling the aforementioned problems with a search algorithm encompassing multiple objectives, including uncertainty, diversity, and weighted coverage. The uncertainty objective helps us find the most informative samples, while the diversity objective reduces the number of similar samples. Coverage objectives can increase the chance of selecting a minority class by assigning a larger weight.

## 8 THREATS TO VALIDITY

**Construct Validity** is the extent to which a metric assesses the theoretical construct it is intended to measure. In this paper, we construct the CCDT to perform a multi-label classification and measure its effectiveness with three commonly used classification metrics, i.e., precision, recall, and F1 score. However, there exist other classification metrics, such as accuracy. Regardless of the popularity of the accuracy metric in classification tasks, we argue that accuracy can be misleading in some cases, especially with imbalanced datasets [5]. By contrast, precision, recall, and F1 score still provide meaningful insights even with an imbalanced dataset. In our context, we calculate precision, recall, and F1 score for each result code $y \in$ {"info", "warning", "not applied", "error"}. Precision measures the percentage of correctly classified cancer messages among all cancer messages predicted as $y$; recall is the percentage of correctly classified cancer messages among all cancer messages with result code $y$; F1 score is a harmonic average of precision and recall, comprehensively reflecting the model performance.

**Internal Validity** is the extent to which the experiments can establish a causal relationship between the independent and dependent variables. In our context, we attempt to demonstrate a causal relationship between the employment of TL and performance improvement. One possible threat to internal validity resides in the selection of hyperparameters. The good performance of a method can come from extensive work on manually selecting an optimal set of hyperparameters, which incorporate external domain knowledge into the model. In this paper, however, we acquire the hyperparameters through experiments. Concretely, we reserve a validation dataset and explore different combinations of hyperparameter values. We choose the best combination as our hyperparameters. Such a process requires no domain knowledge, reducing potential threats to internal validity.

**Conclusion Validity** concerns the statistical significance of the experiment results. The employment of neural networks inherently introduces randomness into the model. In other words, the improvement presented in the experiment can be random and not reproducible. To reduce the influence of randomness, we repeat all the experiments ten times and perform statistical testing to demonstrate the significance of each improvement.

**External Validity** is the extent to which the proposed method can be generalized to other datasets and domains. In this paper, we address external threats from both approach design and dataset construction. First, we design EvoCLINICAL as a generic method, making no strong assumptions on the dataset distribution. Thus, it is intrinsically straightforward to apply EvoCLINICAL in another dataset. Second, we construct three target GURI with evolutions of different difficulty levels in terms of the number of message fields involved in an individual rule. The dataset collected from such diverse evolutions is more representative, further reducing the threat to external validity.

## 9 CONCLUSION AND FUTURE WORK

We propose EvoCLINICAL to construct and evolve CCDT of GURI by leveraging active transfer learning, which fine-tunes a pretrained CCDT with labelled data acquired from the target GURI. The active learning paradigm reduces the query times on the target GURI. To evaluate EvoCLINICAL, we obtained three transfer learning experiment groups and performed comparative experiments with EvoCLINICAL and the baselines. The experimental results demonstrate the effectiveness of EvoCLINICAL with precision, recall and F1 score as at least 0.9200, 0.9105 and 0.9137, respectively. We further demonstrated that active learning contributes to the overall performance of EvoCLINICAL.

In the future, we plan to explore EvoCLINICAL on other datasets in the healthcare domain, such as healthcare information system platforms. For instance, we can build a CCDT for such platforms and adopt EvoCLINICAL to evolve it as the platforms evolve. We will also investigate other design options for EvoCLINICAL, such as long short-term memory and transformer models to deal with time-series data.

## 10 ACKNOWLEDGEMENTS

This work is supported by the AIT4CR project (No. #309642) funded by the Research Council of Norway. C. Lu is supported by the Co-evolver project (No. #286898/F20) funded by the research council of Norway. The experiment has benefited from the Experimental Infrastructure for Exploration of Exascale Computing (eX3), which is financially supported by the Research Council of Norway under contract 270053.